\documentclass[runningheads]{llncs}

\usepackage{amsmath}
\usepackage{multirow}
\usepackage[caption=false]{subfig}
\usepackage{enumitem}
\usepackage{color}
\usepackage{graphicx}
\usepackage{orcidlink}

\begin{document}

\title{Slicing Through Bias: Explaining Performance Gaps in Medical Image Analysis using Slice Discovery Methods}

\titlerunning{Slicing Through Bias}

\author{Vincent Olesen\inst{1}\and
Nina Weng\inst{1}\orcidlink{0009-0006-4635-0438} \and
Aasa Feragen\inst{1}\orcidlink{0000-0002-9945-981X} \and
Eike Petersen\inst{1,2}\orcidlink{0000-0003-0097-3868}}
\authorrunning{Olesen et al.}

\institute{Technical University of Denmark, Kongens Lyngby, Denmark\\
\email{\{ninwe, afhar, ewipe\}@dtu.dk}\\
\and
Fraunhofer Institute for Digital Medicine MEVIS, Bremen, Germany
}
\maketitle             

\begin{abstract}
Machine learning models have achieved high overall accuracy in medical image analysis.
However, performance disparities on specific patient groups pose challenges to their clinical utility, safety, and fairness.
This can affect known patient groups -- such as those based on sex, age, or disease subtype -- as well as previously unknown and unlabeled groups.
Furthermore, the root cause of such observed performance disparities is often challenging to uncover, hindering mitigation efforts.
In this paper, to address these issues, we leverage Slice Discovery Methods (SDMs) to identify interpretable underperforming subsets of data and formulate hypotheses regarding the cause of observed performance disparities.
We introduce a novel SDM and apply it in a case study on the classification of pneumothorax and atelectasis from chest x-rays. 
Our study demonstrates the effectiveness of SDMs in hypothesis formulation and yields an explanation of previously observed but unexplained performance disparities between male and female patients in widely used chest X-ray datasets and models. 
Our findings indicate shortcut learning in both classification tasks, through the presence of chest drains and ECG wires, respectively. Sex-based differences in the prevalence of these shortcut features appear to cause the observed classification performance gap, representing a previously underappreciated interaction between shortcut learning and model fairness analyses.

\end{abstract}

\section{Introduction}

Machine learning models have shown great promise in medical image-based diagnosis, sometimes with performance claims that rival human experts. 
However, reported performance may overstate these models' clinical utility and safety~\cite{Wynants2020}.
Specifically, models may underperform or fail systematically on critical subsets of data even while overall average accuracy remains high. 
In computer vision research, such subsets are called \emph{underperforming slices} or \emph{blind spots}
\cite{eyuboglu_domino_2022,plumb_towards_2023}, where `slice' refers to a subset of samples with similar characteristics, such as an attribute familiar to a domain expert.\footnote[1]{In the medical imaging literature, the term `slice' commonly refers to a two-dimensional cross-section within three-dimensional volumetric data. We are adopting a differing terminology from earlier work on SDMs originating outside of the medical image analysis field. We apologize for the unfortunate clash of terminology.}

In medical image analysis, a model might be underperforming on a slice of patients for a wide range of reasons, including group under-representation, increased input or label noise, fundamental differences in the difficulty of the prediction task, and shortcut learning~\cite{degrave_ai_2021,jimenez-sanchez_detecting_2023,oakden-rayner_hidden_2020,Petersen2023a}.
Performance disparities between patient groups have been observed in many medical imaging domains~\cite{Daneshjou2022,Glocker2023a,Lin2023,seyyed-kalantari_underdiagnosis_2021,Zong2023}, raising concerns about the potential unfairness resulting from the application of such models.
However, properly \emph{mitigating} such performance disparities requires identifying their root cause, which is often challenging~\cite{Mukherjee2022,Petersen2023a,weng_are_2023}.
The challenge is further compounded by the fact that the feature that causally distinguishes high-performing from low-performing patients is often unknown and, thus, not annotated.
This renders simple subgroup analyses based on available metadata insufficient for identifying the causes of performance disparities.

\begin{figure}[t]
    \centering
    \includegraphics[width=\textwidth]{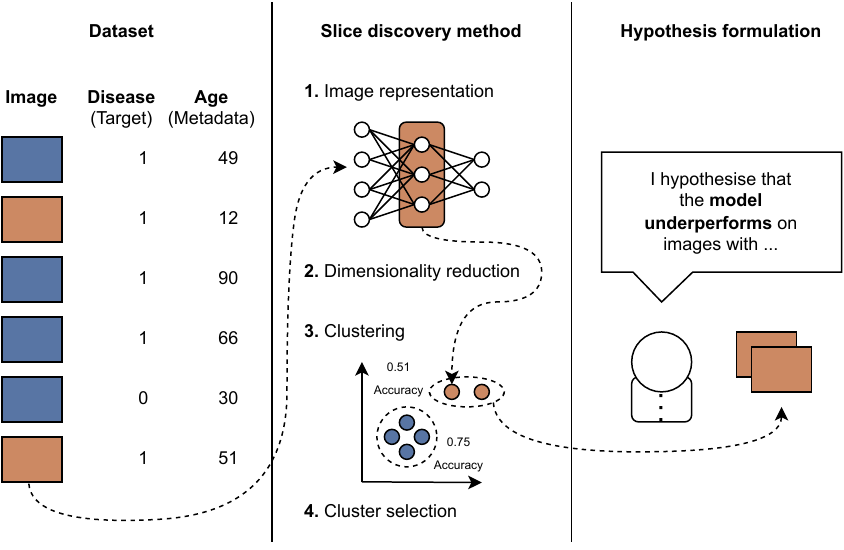}
    \caption{A general overview of the key elements of slice discovery methods.}
    \label{fig:sdm-framework}
\end{figure}
To address the issue of unknown distinguishing features, various methods for the unsupervised discovery of underperforming slices have been proposed in the computer vision literature.
Such methods are variously known as \emph{Slice Discovery Methods} (SDMs)~\cite{eyuboglu_domino_2022} or \emph{Blindspot Discovery Methods} (BDMs)~\cite{plumb_towards_2023}.
Typically, these methods perform a cluster analysis on the input space and then select poorly performing clusters, or \emph{slices} of data, for further analysis; refer to Fig.~\ref{fig:sdm-framework} for a high-level overview.
SDMs can aid machine learning practitioners and domain experts in identifying underperforming sets of data, as well as in forming hypotheses about the \emph{causes} of this underperformance.
With few exceptions~\cite{oakden-rayner_hidden_2020}, SDMs have not yet found widespread use in the medical imaging domain.

In this study, we explore the use of SDMs for the analysis of performance disparities in medical image analysis.
Our contributions are twofold.
First, we provide a general overview of SDMs in medical image analysis and we propose a novel SDM, rigorously justifying all of our design choices.
We demonstrate the effectiveness of our proposed SDM for hypothesis formulation in a case study of pneumothorax and atelectasis classification on two public chest X-ray datasets (NIH-CXR~14~\cite{wang_chestx-ray8_2017} and CheXpert~\cite{irvin_chexpert_2019}).
Second, by further analyzing the hypotheses generated using our SDM, we show that chest drain shortcut learning causes a previously unexplained yet variously reproduced performance gap between male and female subjects in pneumothorax classification.
This constitutes an important link between shortcut learning and model fairness analyses that has, to the authors' knowledge, not been described before.
In addition, using our SDM, we discover a new shortcut feature (the presence of ECG cables) that may explain male--female performance disparities in atelectasis classification.

\section{Related work}
\subsection{Bias and shortcuts in chest x-ray analysis}
Algorithmic fairness in medical image analysis, and performance disparities between patient groups in particular, have recently come under rapidly increasing scrutiny~\cite{Daneshjou2022,larrazabal_gender_2020,oakden-rayner_hidden_2020,RicciLara2023,seyyed-kalantari_underdiagnosis_2021,weng_are_2023}.  
In this context, the fairness of chest x-ray-based disease classification models has received particularly broad attention~\cite{Brown2023,Glocker2023a,larrazabal_gender_2020,seyyed-kalantari_underdiagnosis_2021,weng_are_2023,Zhang2022}. 
Larrazabal et al.~\cite{larrazabal_gender_2020} demonstrated that such models had better classification performance for a particular patient group (based on biological sex) if that group was represented in higher proportions of the training dataset.
While not the focus of their study, their results also indicated significant differences between model performance on male and female subjects, with the classification models performing better for either group in different diseases.
These (sometimes large) performance gaps persisted even in the case of sex-balanced training sets.
This observation prompted Weng et al.~\cite{weng_are_2023} to investigate the hypothesis that biological sex differences were causing these unexplained performance gaps.
Based on their results, the authors dismissed breast shadows as a factor, but other biological sex differences contributing to performance gaps remain uncertain, leading to an unexplained gender disparity.
Zhang et al.~\cite{Zhang2022} employed standard algorithmic fairness mitigation approaches to the chest x-ray case, finding that simple group balancing was one of the most robust approaches -- which did not, however, mitigate the performance gaps observed by Larrazabal et al.~\cite{larrazabal_gender_2020}.

In a separate development, it has been widely demonstrated that chest x-ray-based disease classification models are prone to relying on shortcut learning~\cite{Brown2023,degrave_ai_2021,jimenez-sanchez_detecting_2023,oakden-rayner_hidden_2020}. 
Both Oakden-Rayner et al.~\cite{oakden-rayner_hidden_2020} and Jiménez-Sánchez et al.~\cite{jimenez-sanchez_detecting_2023} demonstrate how pneumothorax classification tends to rely heavily on the presence of chest drains, which represent the standard treatment for pneumothorax.
Connecting the two challenges of shortcut learning and fairness, many authors have raised concerns about the potential for deep learning models to exploit spurious correlations between sensitive attributes, such as age, gender, or ethnicity, and the prediction target~\cite{Brown2023,Glocker2023a}.  
To the authors' knowledge, the fact that shortcut learning relying on \emph{non-}sensitive features (such as the presence of chest drains) can explain performance disparities between sensitive groups (such as gender groups) has not been discussed explicitly before.
Notably, Jiménez-Sánchez et al.~\cite{jimenez-sanchez_detecting_2023} took important first steps in this direction, by differentially reporting the effect of shortcut learning on different gender groups.

\subsection{Slice discovery methods}\label{sec:related-work-sdm}
Slice discovery methods (SDMs) are a recently emerging tool for the performance analysis and subsequent improvement of deep learning models.
In particular, they aim to solve the problem that the features that identify underperforming groups of inputs might not be known a priori.
To this end, SDMs typically perform unsupervised clustering of the input data, in order to identify semantically similar `slices' of data that the model under analysis performs poorly on.
In more detail, SDMs usually consist of the following steps: (1) the input data is embedded into a latent representation space, (2) some SDMs perform dimensionality reduction, (3) an unsupervised learning method, such as clustering, is used to extract slices of the data, and (4) the extracted slices are prioritized based on a performance metric, such as accuracy.
In table~\ref{tab:sdm-related-work}, adapted from Plumb et al.~\cite{plumb_towards_2023}, we summarize previously proposed SDMs and their respective high-level design.
\begin{table}[t]
\centering
\caption{A summary of slice discovery methods.
Clf: The representation used by the classification model under analysis.
Adapted from Plumb et al.~\cite{plumb_towards_2023}.}
\label{tab:sdm-related-work}
\begin{tabular}{llll}
\hline
\textbf{Method} & \textbf{Rep.} & \textbf{Dim. reduction} & \textbf{Clustering} \\ \hline
Algorithmic measurement \cite{oakden-rayner_hidden_2020} & Clf &  & KNN \\
\multirow{ 2}{*}{MultiAccuracy Boost \cite{kim_multiaccuracy_2019}} & \multirow{ 2}{*}{VAE} &  & Rigid/decision-tree \\
&&& regression\\
GEORGE \cite{sohoni_no_2022} & Clf/BiT emb. & UMAP (d=1,2) & GMM \\
Spotlight \cite{deon_spotlight_2021} & Clf &  & Optimization problem \\
Planespot \cite{plumb_towards_2023} & Clf & scvis (d=2) & GMM \\
Domino \cite{eyuboglu_domino_2022} & CLIP & PCA (d=128) & GMM \\
Failure mode distillation \cite{jain_distilling_2022} & CLIP &  & SVM \\ 
Bias-Aware Hierarchical  & \multirow{ 2}{*}{Clf}   & \multirow{ 2}{*}{UMAP(d=2)} & \multirow{ 2}{*}{Mod. K-Means} \\
Clustering~\cite{MisztalRadecka2021} & &\\
\hline
\textbf{Proposed SDM (Ours)} & Clf & FC layer (d=128) & GMM \\ \hline
\end{tabular}
\end{table}
The most common methods to extract slices are clustering algorithms, namely the Gaussian Mixture Model (GMM) and its variants.
The most common choice of image embedding is the latent space representation computed by the classification model under scrutiny, i.e., its penultimate layer's outputs.
However, recent methods instead utilize multi-modal pre-trained models like CLIP to enable the generation of text descriptions for extracted slices~\cite{eyuboglu_domino_2022,jain_distilling_2022}.
Interestingly, the crucial dimensionality reduction step has received relatively little attention yet, as recently pointed out by Plumb et al.~\cite{plumb_towards_2023}.
Possibly due to their relatively recent emergence, SDMs have not yet been widely applied in the medical image domain.
In this regard, the work of Oakden-Rayner et al.~\cite{oakden-rayner_hidden_2020} represents a very notable early exception that precedes more recent SDM developments.

\section{Methodology}
\subsection{Proposed slice discovery method}
This section introduces our proposed SDM and motivates our design choices.
The proposed method consists of the following four steps, following Fig.~\ref{fig:sdm-framework}:
\begin{description}
\item[Image representation.]{
We use the image representation computed by the penultimate layer of the classification model under scrutiny.
As opposed to SDMs that use a separate model to obtain image embeddings, our approach relies only on information available to the model’s final classification layer.} 
\item[Dimensionality reduction.]{
We insert a single fully connected layer with d-dimensional output (and a sigmoid activation layer)   between the classification model's penultimate layer and its final classification head. We train (just) this additional layer following the same procedure that was used for the classification model itself.
Similarly to the previous step and contrary to standard choices such as PCA, t-SNE, or UMAP, our approach is \emph{supervised} and preferably preserves information relevant to the model's predictions.}
\item[Clustering.]{
We use a Gaussian Mixture Model (GMM) for clustering and the Bayesian Information Criterion (BIC) for choosing the number of clusters.
We cluster disease-positive and -negative samples separately to extract clusters of the same error type, similar to Oakden-Rayner et al.~\cite{oakden-rayner_hidden_2020}. 
}
\item[Cluster selection.]{We propose using the Brier score (BS), a proper scoring rule, to prioritize under- and overperforming slices, equivalent to mean squared error between model confidence and binary target labels. 
The main motivation for our choice is that, as opposed to classification accuracy, the BS is threshold-independent. 
In addition, it captures both the model's discriminative ability and its calibration~\cite{Broecker2009}.
As opposed to AUROC~\cite{Kallus2019}, per-cluster BS can be meaningfully compared between groups, and as opposed to many calibration metrics~\cite{Petersen2023,RicciLara2023}, it can be meaningfully compared between clusters of different sizes.
We quantify BS uncertainty by simple bootstrapping of each cluster and select the cluster with the lowest 97.5-quantile as the best, and the cluster with the highest 2.5-quantile as the worst.
}
\end{description}

\subsection{Datasets}
We consider a case study on two public datasets, NIH-CXR14~\cite{wang_chestx-ray8_2017}
and CheXpert~\cite{irvin_chexpert_2019}.
Both datasets slightly over-represent male subjects.
We reused chest drain labels previously crowd-sourced from both radiologists and nonexperts \cite{damgaard_augmenting_2023,oakden-rayner_hidden_2020,jimenez-sanchez_detecting_2023}, including 3543 random cases with pneumothorax in the NIH dataset and 972 cases with and without pneumothorax in CheXpert.
For NIH, we observe a larger prevalence of chest drains among pneumothorax-positive male subjects compared to female subjects (49.5\% vs. 42.8\%).
For CheXpert, we observe a larger prevalence of chest drains among pneumothorax-negative male subjects (23.0\% vs. 14.7\% in females) but comparable chest drain prevalence across sexes for pneumothorax-positive subjects (50.5\% in males vs. 50.2\% in females).

\subsection{Experiments}
We conduct a case study on the pneumothorax classification task, following the experimental setup of Weng et al.~\cite{weng_are_2023}.
Specifically, to reduce the potential for label noise to affect our analyses, we select one sample per patient, with a preference for pneumothorax-positive samples and an equal sex ratio.
Withholding the chest drain-annotated samples, we split the datasets into 60\%/10\%/30\% train, validation, and test sets, resampling the splits ten times.
We resize the images to 224x224 pixels and train a ResNet50 (Adam optimizer, learning rate $10^{-6}$, batch size 64, 20 epochs).
We use data augmentation for the training dataset, including horizontal flipping, rotation up to 15 degrees, and scaling up to 10\% with a 50\% probability for each augmentation.
We then carry out our proposed SDM and report the distribution of comorbidities and chest drains (for annotated chest drain samples).\footnote[2]{For the dimensionality reduction, we used d=128 following similar previous work~\cite{eyuboglu_domino_2022}. Results with d=10 were comparable. The gap statistic~\cite{Tibshirani2001} indicates that clustering does indeed occur in the reduced space.
The effect of the additional dimensionality reduction layer on the model's classification performance was negligible in terms of test accuracy and AUROC.}
We repeat the same analyses for atelectasis classification.
Based on our findings, we conduct further post-hoc analyses in both cases. 
Our source code is publicly available at \href{https://github.com/volesen/slicing-through-bias}{https://github.com/volesen/slicing-through-bias}. 

\section{Results}
\begin{figure}[t!]
    \centering
         \includegraphics[width=\textwidth]{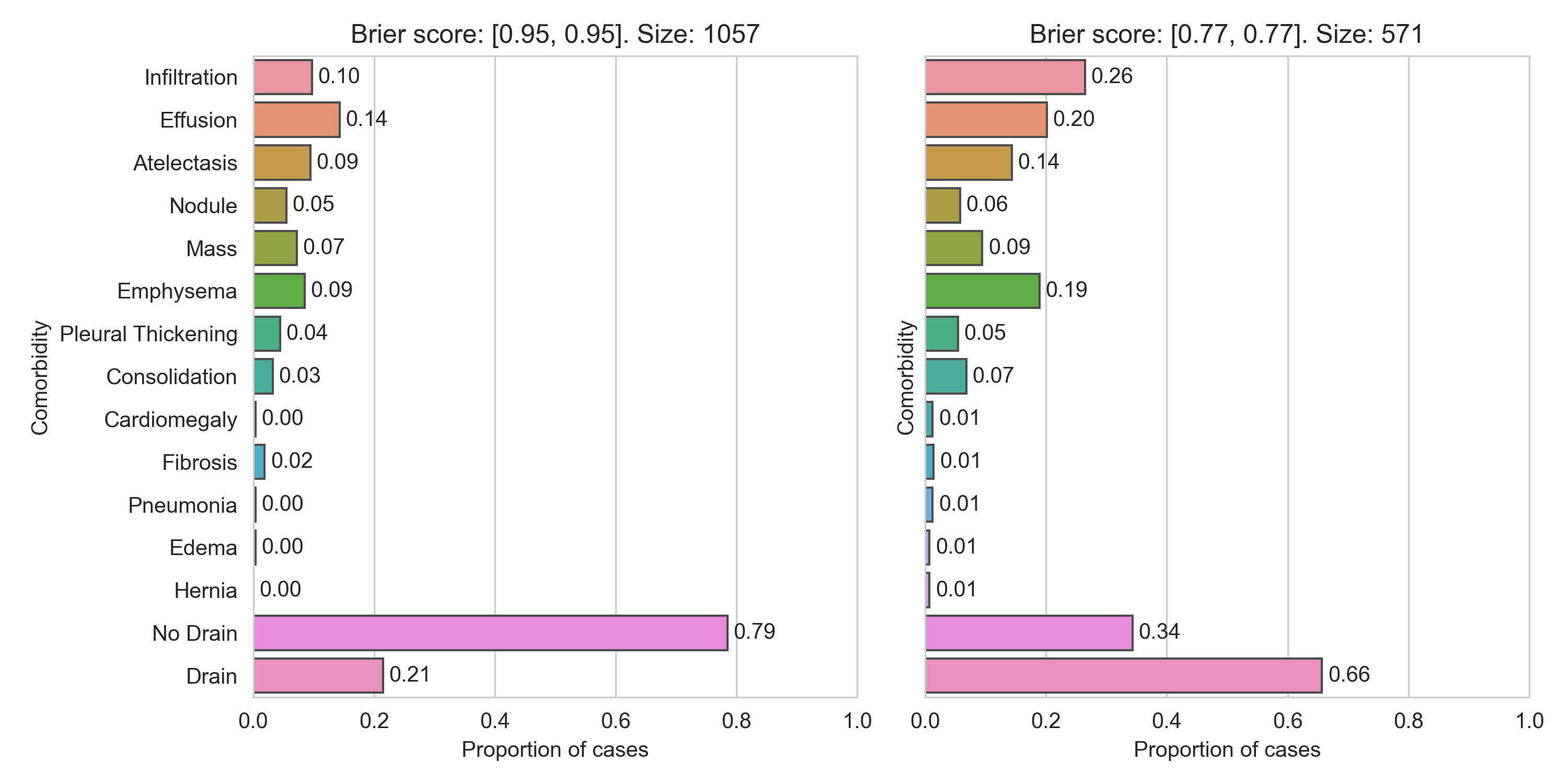}
         \label{fig:nih-pneumothorax-positive-comorbidities}
    \caption{The comorbidity and chest drain distribution in pneumothorax-positive chest drain annotated samples of NIH-CXR14 for the worst-performing (left column) and best-performing (right column) slices by Brier score.
    The pneumothorax-negative case is omitted as chest drain annotations were not available for these samples in NIH-CXR14.
    }
    \label{fig:nih-pneumothorax-comorbidities}
\end{figure}
\begin{figure}[h]
    \centering
    \subfloat[CheXpert, Pneumothorax Negative\label{fig:chexpert-pneumothorax-negative-comorbidities}]{
         \centering
         \includegraphics[width=0.85\textwidth]{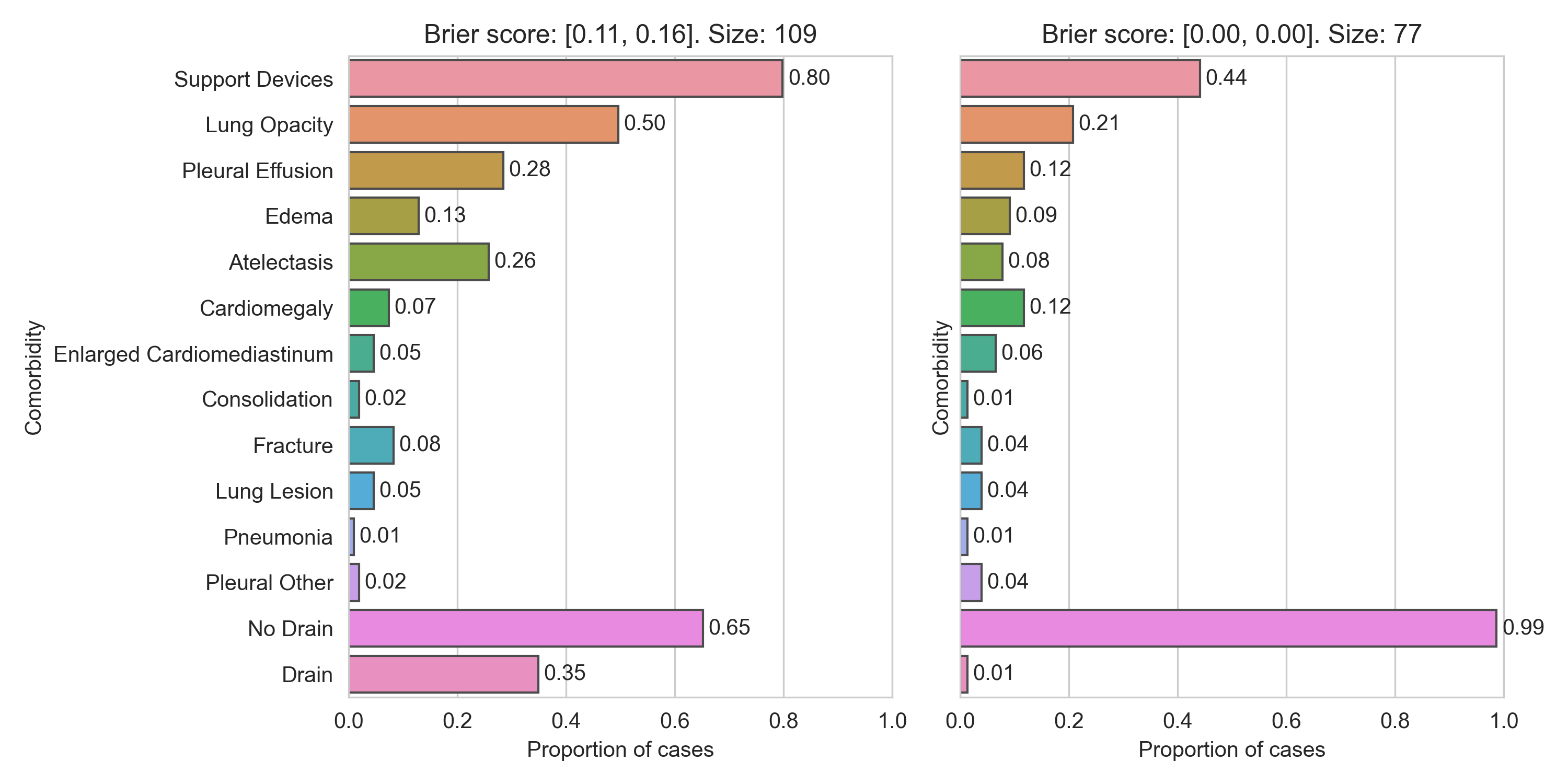}
    }
    \hfill
    \subfloat[CheXpert, Pneumothorax Positive\label{fig:chexpert-pneumothorax-positive-comorbidities}]{
         \centering
         \includegraphics[width=0.85\textwidth]{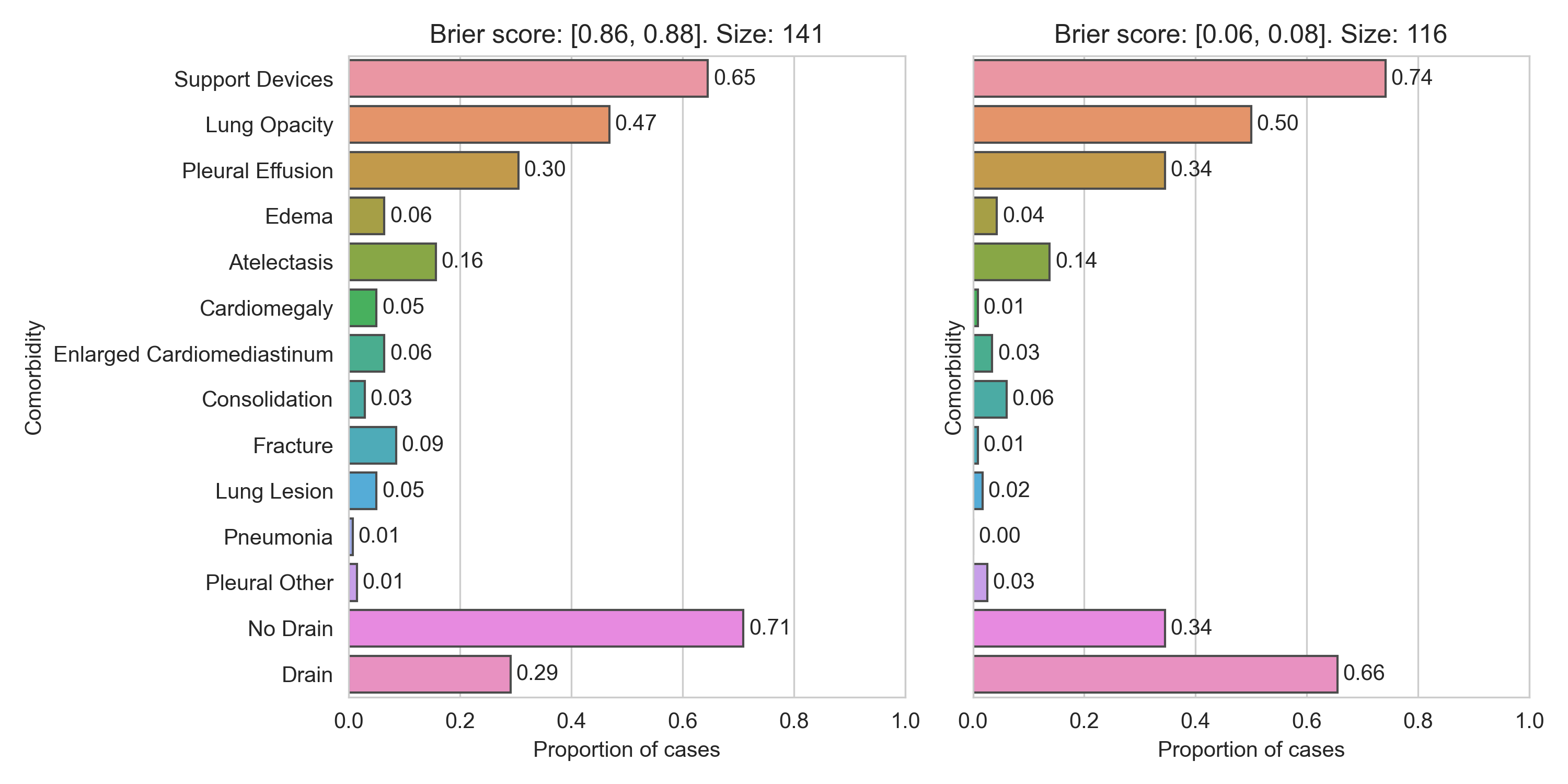}%
    }
    \caption{The comorbidity and chest drain distribution in (a) pneumothorax-negative (top row) and (b) pneumothorax-positive chest drain annotated samples of CheXpert for the worst-performing (left column) and best-performing (right column) slices by upper 95\% bootstrapped Brier scores.
    }
    \label{fig:chexpert-pneumothorax-comorbidities}
\end{figure}

\paragraph{Pneumothorax classification.}
Fig.~\ref{fig:nih-pneumothorax-comorbidities} and fig.~\ref{fig:chexpert-pneumothorax-comorbidities} show the best- and worst-performing slices based on the Bier score 
in the NIH-CXR14 and CheXpert datasets.
In both datasets, the underperforming slices for pneumothorax-negative samples have a lower-than-average chest drain proportion, while the opposite holds for pneumothorax-positive cases.

\begin{figure}[t]
    \centering
         \includegraphics[width=0.6\textwidth]{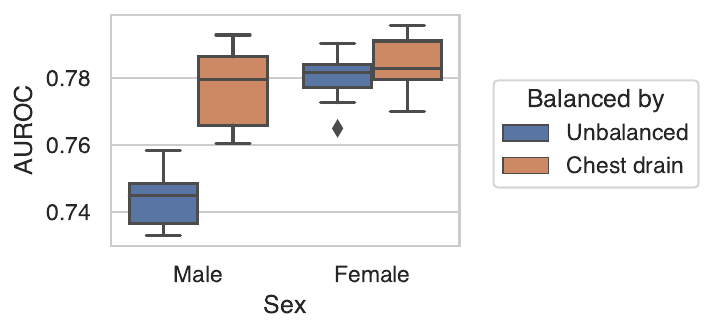}
    \caption{
    AUROC on CheXpert with male and female test subjects on pneumothorax prediction, following the natural (`unbalanced') distribution of chest drains and balanced by chest drain presence across ten samplings of the train-validation-test sets.}
    \label{fig:unbalanced-vs-balanced}
\end{figure}
\begin{figure}[h]
    \centering
    \subfloat[NIH-CXR14\label{fig:nih-confidence-by-drain}]{
         \centering
         \includegraphics[width=0.33\textwidth]{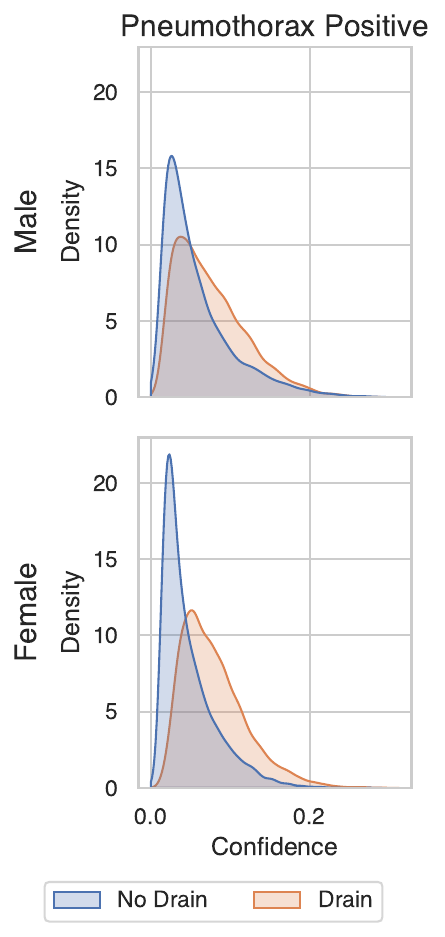}%
    }
    \hfill
    \subfloat[CheXpert\label{fig:chexpert-confidence-by-drain}]{
        \centering
        \includegraphics[width=0.65\textwidth]{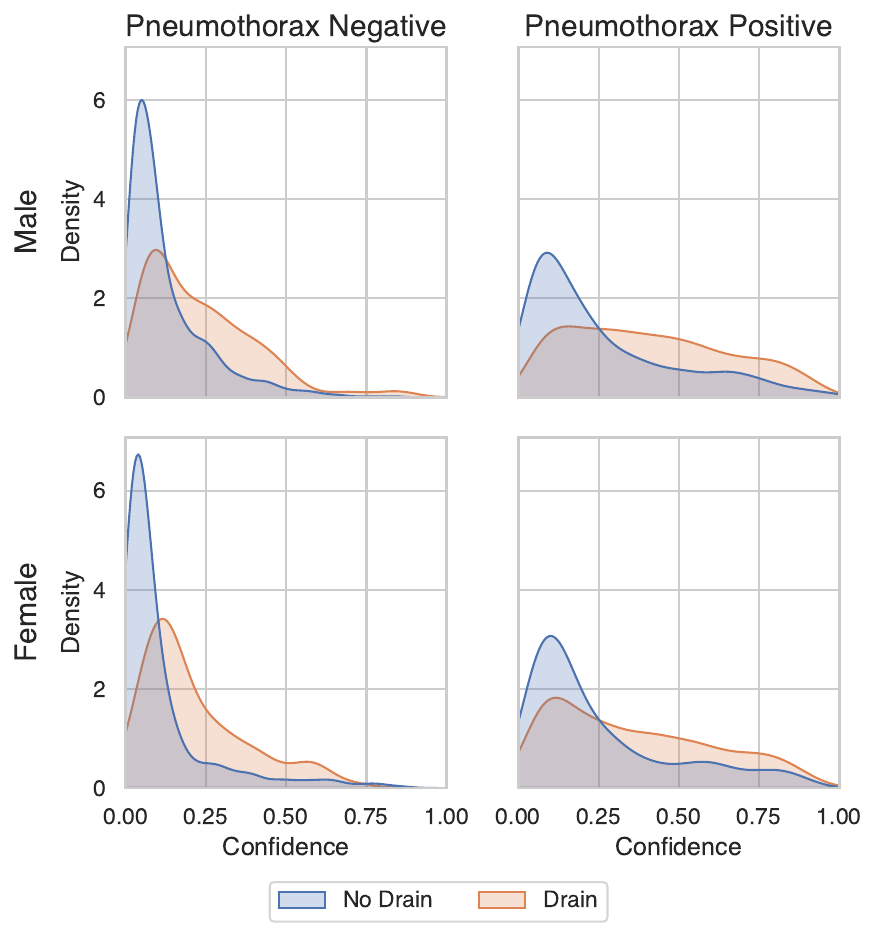}%
    }
    \caption{Distribution of confidences (the softmax output of the model for the disease-positive class) for pneumothorax classification by sex, presence of pneumothorax, and chest drain for NIH-CXR14 (left) and CheXpert (right).
    Throughout, subjects without chest drains are more likely to be classified as pneumothorax-negative.}
    \label{fig:confidence-by-drain}
\end{figure}

Based on these results (and based on prior work~\cite{jimenez-sanchez_detecting_2023,oakden-rayner_hidden_2020}), we hypothesized that the presence of chest drains is used to classify pneumothorax. 
Indeed, we observe that computed model confidence (the softmax output of the model for the disease-positive class) is consistently higher in samples with chest drains across datasets, sexes, and pneumothorax labels (positive/negative), lending strong support to this hypothesis (fig.~\ref{fig:confidence-by-drain}).
Furthermore, as both datasets have varying chest drain prevalences by sex, we hypothesized that this could contribute to the gender performance gap.
To assess this hypothesis, we determined AUROC by sex, firstly in a test set following the natural distribution of chest drains, and secondly a test set with equalized chest drain prevalence in the male and female populations (Fig.~\ref{fig:unbalanced-vs-balanced}).
We observe a significant difference in performance in the first case ($p < 0.001$) but no significant difference in the second, chest-drain-balanced case ($p > 0.1$), indicating that chest drain shortcut learning is the cause of a large part of the male--female performance gap observed in prior work~\cite{larrazabal_gender_2020,weng_are_2023}.
Statistical significance was assessed using a Mann-Whitney U-test.
It should be noted that the differences in comorbidities noticeable in Fig.~\ref{fig:nih-pneumothorax-comorbidities} represent another potential explanation for the observed performance disparities. While we did not further explore this hypothesis here, our additional experiments related to the chest drain shortcut hypothesis (discussed above) seem to indicate that this indeed explains the bulk of the observed performance gaps.

\paragraph{Atelectasis classification.}
We repeated our SDM analysis for atelectasis.
(Slice statistics not shown here due to space constraints.)
Prompted by a visual inspection of the recordings in the best and worst-performing slices, we hypothesized that the two slices differed substantially in their prevalence of ECG cables in the recordings.
To test this hypothesis, we randomly selected and (as non-experts) labeled 100 samples each for the best- and worst-performing slices on atelectasis-positive and -negative cases according to whether they displayed ECG cables or not. 
Of atelectasis-negative samples, 95\% of the labeled recordings in the worst-performing slice contained ECG cables, compared to 10\% in the best-performing slice. 
Of the atelectasis-positive cases, 50\% of the labeled recordings in the worst-performing slice had ECG cables, compared to 99\% in the best-performing slice.

\section{Discussion and conclusion}
We have proposed a novel slice discovery method (SDM), which differs from previously proposed methods in several key elements.
Both in the representation extraction as well as in the (supervised) dimensionality reduction step, we prioritize only using information available to the classification model under test.
In addition, we propose using the Brier score (BS) for selecting highly and poorly performing clusters because it is threshold-independent, accounts for both discriminative ability and calibration, and enables meaningful comparisons between clusters of different sizes.
In a case study on chest x-ray-based disease classification, our SDM successfully recovered a previously known case of shortcut learning (chest drains for pneumothorax classification) and suggested a new, previously unknown case (ECG cables for atelectasis classification). 
The latter case also demonstrated another benefit of using SDMs: reducing the required labeling efforts, because these can be specifically targeted at the worst and best clusters. 
Our case study shows that our proposed SDM, and SDMs in general, can aid researchers in generating hypotheses regarding the causes of model underperformance on subsets of data, which is crucial for leveling \emph{up} performance~\cite{Petersen2023a}.

Consistent with the observations made by Weng et al.~\cite{weng_are_2023}, our findings challenge the notion that biological differences are the primary driver of the previously observed~\cite{larrazabal_gender_2020} but unexplained male--female performance gaps in chest x-ray-based disease classification.
Instead, our results suggest that shortcut learning in conjunction with a difference in chest drain prevalence between males and females causes the observed performance disparity.
This newly gained knowledge opens up the possibility for the targeted application of shortcut learning mitigation techniques, instead of relying on blind group performance equalization approaches that often result in leveling \emph{down} performance~\cite{Petersen2023a,Zhang2022,zietlow_leveling_2022}.

While our study builds upon the chest drain annotations of  \cite{damgaard_augmenting_2023} and \cite{jimenez-sanchez_detecting_2023}, for which the former show a high level of agreement between expert and non-expert annotations, caution is warranted. Non-expert annotations, although showing agreement, may not necessarily represent the ground truth or offer a representative sample (images without consensus between multiple labelers were disregarded~\cite{jimenez-sanchez_detecting_2023}). 
This caution is emphasized by a significant difference in the prevalence of chest drains among pneumothorax-positive samples in NIH-CXR14 between studies~\cite{damgaard_augmenting_2023,oakden-rayner_hidden_2020}. 
Moreover, the mitigation of shortcut learning is a highly active research area, and mitigating many different shortcuts simultaneously, and with limited label availability, remains challenging~\cite{Li2023a}.
Finally, the interpretation of identified slices in more challenging cases represents a crucial challenge for future research. Both chest drains and ECG cables are visible to the human eye (though maybe not the non-expert), but other potentially problematic features may not be.

\subsection*{Acknowledgements}
Work on this project was partially funded by the Independent Research Fund Denmark (DFF, grant number 9131-00097B), Denmark’s Pioneer Centre for AI (DNRF grant number P1), and the Novo Nordisk Foundation through the Center for Basic Machine Learning Research in Life Science (MLLS, grant number NNF20OC0062606). The funding agencies had no influence on the writing of the manuscript nor on the decision to submit it for publication.

\begin{figure}[h]
    \centering
    \subfloat[NIH-CXR14\label{fig:nih-Brier-score-by-drain}]{%
         \centering
         \includegraphics[width=0.3\textwidth]{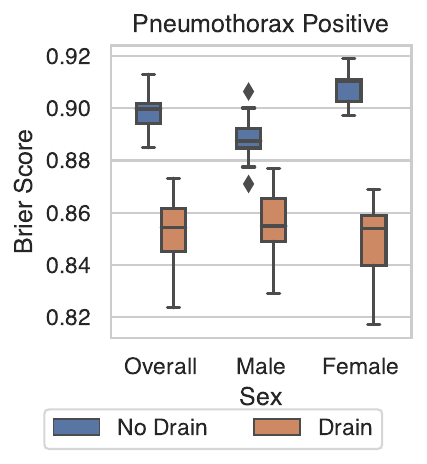}%
    }
    \hfill
    \subfloat[CheXpert\label{fig:chexpert-Brier-score-by-drain}]{%
        \centering%
        \includegraphics[width=0.6\textwidth]{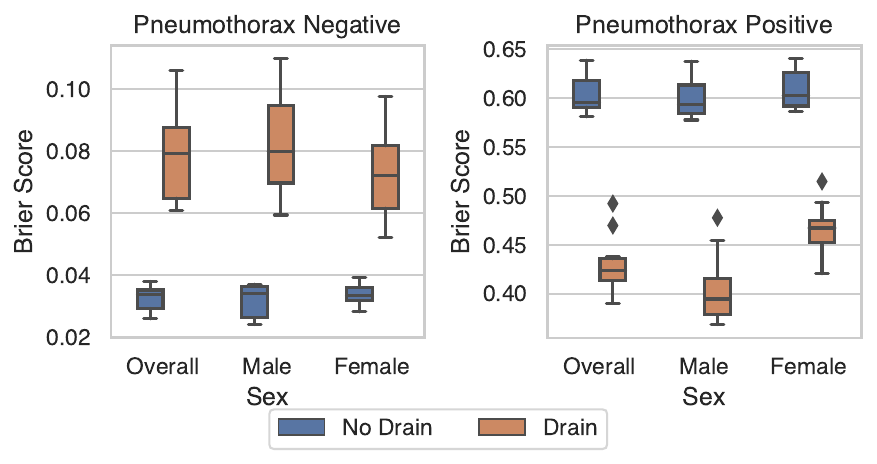}%
    }
    \caption{Brier scores for pneumothorax classification 
    for NIH-CXR14 (left) and CheXpert (right) across ten samplings of the train-validation-test sets. 
    Throughout, non-pneumothorax patients with chest drains and pneumothorax patients without chest drains are underperforming compared to pneumothorax-positive subjects with chest drains and pneumothorax-negative subjects without chest drains.
    }
    \label{fig:Brier-score-by-drain}
\end{figure}

\bibliographystyle{splncs04}
\bibliography{citations}

\end{document}